\documentclass[11pt, final, letterpaper,oneside,onecolumn]{IEEEtran}
\usepackage[colorlinks=true,citecolor=magenta,urlcolor=blue]{hyperref}
\usepackage{cite}
\usepackage{graphicx,epstopdf}
\usepackage{rotating}
\usepackage{epsfig}
\usepackage[cmex10]{amsmath}
\usepackage{amssymb}
\usepackage{float}
\usepackage[tight,footnotesize]{subfigure}
\usepackage{array}
\usepackage{color}
\usepackage{multirow}
\ifCLASSINFOpdf
\else
\fi

\begin{document}
%
% paper title
% can use linebreaks \\ within to get better formatting as desired
% Do not put math or special symbols in the title.
\title{Downscaling Microwave Brightness Temperatures Using Self Regularized Regressive Models}

% author names and affiliations
% use a multiple column layout for up to three different
% affiliations
%\author{\IEEEauthorblockN{Subit Chakrabarti}
%\IEEEauthorblockA{Center for \\Computer Engineering\\
%Georgia Institute of Technology\\
%Atlanta, Georgia 30332--0250\\
%Email: http://www.michaelshell.org/contact.html}
%\and
%\IEEEauthorblockN{Homer Simpson}
%\IEEEauthorblockA{Twentieth Century Fox\\
%Springfield, USA\\
%Email: homer@thesimpsons.com}
%\and
%\IEEEauthorblockN{James Kirk\\ and Montgomery Scott}
%\IEEEauthorblockA{Starfleet Academy\\
%San Francisco, California 96678-2391\\
%Telephone: (800) 555--1212\\
%Fax: (888) 555--1212}}

% conference papers do not typically use \thanks and this command
% is locked out in conference mode. If really needed, such as for
% the acknowledgment of grants, issue a \IEEEoverridecommandlockouts
% after \documentclass

% for over three affiliations, or if they all won't fit within the width
% of the page, use this alternative format:
% 
\author{Subit~Chakrabarti,~\IEEEmembership{Student Member,~IEEE}, ~Jasmeet~Judge,~\IEEEmembership{Senior~member,~IEEE},~Anand~Rangarajan{Member,~IEEE},~Sanjay~Ranka{Fellow,~IEEE}\\%
\thanks{Preprint submitted to the IEEE International Geoscience and Remote Sensing Symposium 2015.

This work was supported in part by the NASA-Terrestrial Hydrology Program (THP)-NNX09AK29G and NNX13AD04G.

S.~Chakrabarti and J.~Judge are with the Center for Remote Sensing, Agricultural and Biological Engineering Department, 
Institute of Food and Agricultural Sciences, University of Florida. A.~Rangarajan and S.~Ranka are with the Department of Computer \% Information Science \% Engineering, Gainesville, FL-32611, USA 
E-mail: \href{mailto:subitc@ufl.edu}{subitc@ufl.edu}
}}

% use for special paper notices
%\IEEEspecialpapernotice{(Invited Paper)}

% make the title area
\maketitle

% As a general rule, do not put math, special symbols or citations
% in the abstract
\begin{abstract}
An novel algorithm is proposed to downscale microwave brightness temperatures ($\mathrm{T_B}$), at scales of 10-40 km such as those from the Soil Moisture Active Passive mission to a resolution meaningful for hydrological and agricultural applications. This algorithm, called Self-Regularized Regressive Models (SRRM), uses auxiliary variables correlated to $\mathrm{T_B}$ along-with a limited set of \textit{in-situ} SM observations, which are converted to high resolution $\mathrm{T_B}$ observations using biophysical models. It includes an information-theoretic clustering step based on all auxiliary variables to identify areas of similarity, followed by a kernel regression step that produces downscaled $\mathrm{T_B}$. This was implemented on a multi-scale synthetic data-set over NC-Florida for one year. An RMSE of 5.76~K with standard deviation of 2.8~k was achieved during the vegetated season and an RMSE of 1.2~K with a standard deviation of 0.9~K during periods of no vegetation.
\end{abstract}

% no keywords
\newpage

% For peer review papers, you can put extra information on the cover
% page as needed:
% \ifCLASSOPTIONpeerreview
% \begin{center} \bfseries EDICS Category: 3-BBND \end{center}
% \fi
%
% For peerreview papers, this IEEEtran command inserts a page break and
% creates the second title. It will be ignored for other modes.
\IEEEpeerreviewmaketitle

\section{Introduction}

Spatio-temporal distribution of soil moisture (SM) heavily influences atmospheric and hydrological processes. Accurate SM at spatial scales of 1-5 km is critical for agricultural applications such as drought monitoring, risk management, and productivity predictions, with major implications for food security and sustainability. Microwave observations at frequencies $<$10~GHz are very sensitive to SM in the top 5-10~cm, due to large differences in dielectric constants of dry and wet soils and have been widely used to retrieve SM\cite{Chakrabarti2014, Chakrabarti2013}.

Most downscaling studies have downscaled SM derived from microwave observations~\cite{Merlin2013}, while very few studies have downscaled satellite $\mathrm{T_B}$ observations directly to match model scales. Even for 5 simple soil moisture scenarios with limited variability, and ignoring errors due to model inadequacies, input parameter uncertainties and sensor calibration errors, the resulting error due to sub-footprint variability, was found to be multi-modal and, in some scenarios, had an root mean square error of 0.09 $\mathrm{m^3/m^3}$. Downscaling such a biased product will only increase the final error, even if the downscaling method has built in corrections for physical and meteorological heterogeneity. Thus, downscaling $\mathrm{T_B}$ directly and then assimilating the downscaled product into hydrology models or crop growth models may significantly improve root zone soil moisture and crop yield estimates. Piles et. al.~\cite{Piles2012} downscaled T$_{\textrm{B}}$ directly into SM by applying the class of methods known as Universal Triangle (UT) method to the problem and used a linear regression based linking model to relate low resolution SM to H-Pol and V-Pol T$_{\textrm{B}}$ from the SMOS mission, and other high resolution products, aggregated to the resolution of SMOS observations. Subsequently, using the assumption that the same relationship holds at high resolution, they estimated SM at 1~km. The assumption of scale invariance is not theoretically substantiated and has been found to result in high downscaling errors, particularly during heterogeneous land cover (LC) conditions~\cite{Chakrabarti2014}. Das et. al.~\cite{Das2014} used the correlations between fluctuations of passive radiometer observations and active radar backscatter using a time series approach to obtain a merged SM product at~9~km. Some studies have used statistical inversion techniques like linear inversion with regularization~\cite{YongQian2011}, Singular value decomposition (SVD)~\cite{Gambardella2008} and gradient descent in Banach spaces~\cite{Lenti2014}. A major drawback in these approaches is the assumption of static vegetation conditions and in-footprint spatial homogeneity. Scaling algorithms based upon second-order statistics can potentially lead to significant loss of structural information in the data~\cite{Chakrabarti2013}, especially under highly non-linear heterogeneous and dynamic conditions. Studies involving heterogeneous and dynamic land cover conditions, such as agricultural regions are necessary to understand the validity of downscaling algorithms over real data.
In this study, a novel algorithm is presented that downscales $\mathrm{T_B}$ directly using auxiliary information provided by satellite derived LST, Leaf area index (LAI), precipitation (PPT) and LC at high resolution in conjunction with a very limited set of \textit{in-situ} SM observations. The \textit{in-situ} SM observations are converted into $\mathrm{T_B}$ observations using a biophysical model. In the absence of \textit{in-situ} SM observations, airborne or \textit{in-situ} $\mathrm{T_B}$ observations can also be used for training. In areas without any high resolution data, representative data from other regions with similar topography, slope and soil texture can be used. This algorithm, based on self-regularized regressive models (SRRM) does not make any assumptions about the vegetation conditions and uses higher order statistical descriptors to downscale, and quantify the efficacy of downscaling. The goal of this study is to implement a downscaling algorithm that disaggregates coarse-scale remotely sensed products with auxiliary fine-scale data. The primary objectives are to, 1) estimate $\mathrm{T_B}$ at 1~km using $\mathrm{T_B}$ at 10~km and other spatially correlated variables for a multi-scale synthetic dataset~\cite{Nagarajan2012} based in North-Central Florida, and 2) to conduct a thorough statistical analysis of the downscaled $\mathrm{T_B}$ in order to evaluate the efficacy of the SRRM downscaling algorithm.

\section{Theory}
\label{Sec:Theory}

\begin{figure}
\centering
\includegraphics[width=0.48\textwidth, trim= 2.8cm 5cm 3cm 5cm, clip=true ]{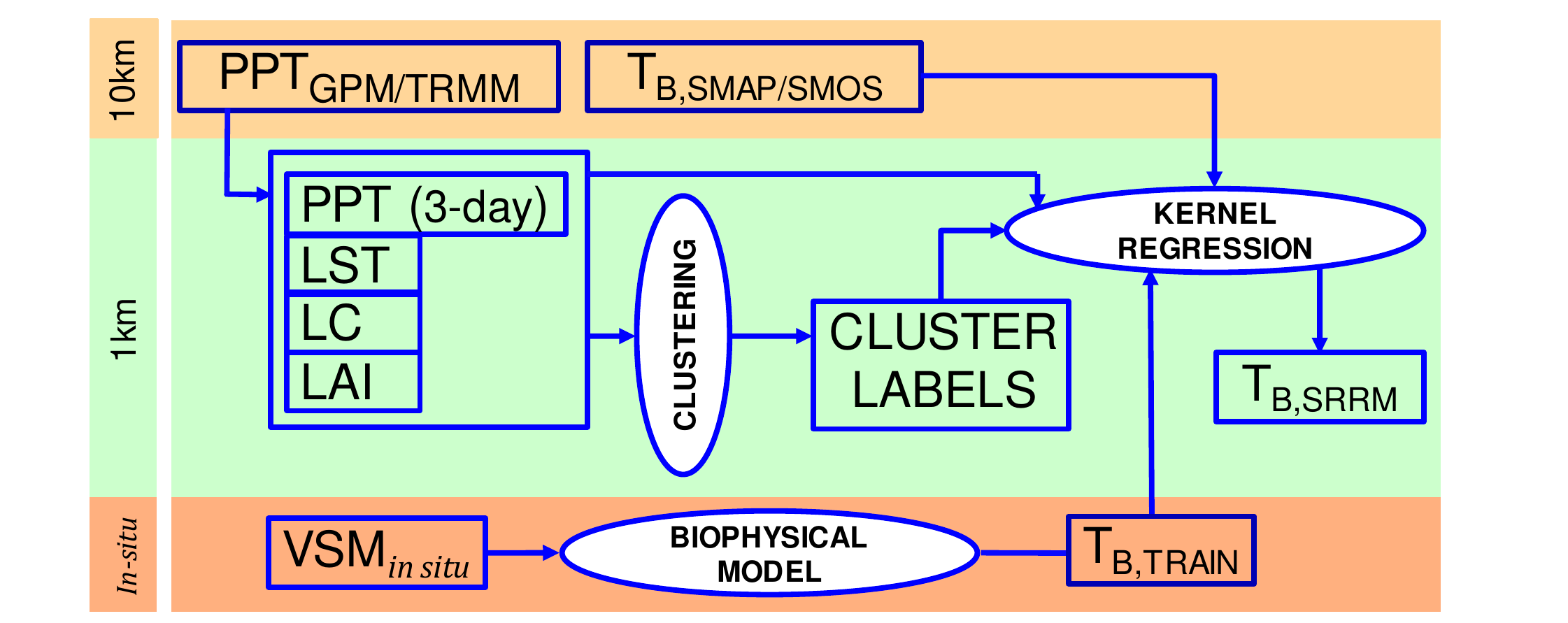}%
\caption{Flowchart of the SRRM algorithm}
\label{fig:flowchart}
\end{figure}

In this study, shown as a flow diagram in Figure~\ref{fig:flowchart}, a number of models are created dynamically based on generalized proximity regions in the high dimensional correlated data. The membership of a pixels to a model is fuzzy, constrained to a sum of one across the model space. The models itself are trained using a kernel regression based method.   The first step of the algorithm, which clusters the study area into proximity regions, is described below. A clustering algorithm which uses information theoretic measures of inter and intra cluster similarity is used~\cite{Jenssen2005}. The membership vectors are not discretized after each step to decrease downscaling errors along cluster boundaries. The cost-function is regularized using the weighted Shannon entropy of the membership vector, such that the membership vectors are sufficiently sparse, as follows.
\begin{align}
\label{eq:JCS_estimator_reg}
\hat{J}_{CS}^{REG} &= \frac{\frac{1}{2} \sum_{i=1}^{N} \sum_{j=1}^{N} \left(1 - \textbf{m}_i^\mathrm{T} \textbf{m}_j\right) G_{\sigma \sqrt{2}} \left(\mathbf{x}_i,\mathbf{x}_j\right)}{\sqrt{\prod_{k=1}^{K} \sum_{i=1}^{N} \sum_{j=1}^{N} m_{ik}m_{jk}G_{\sigma \sqrt{2}}(\mathbf{x}_i,\mathbf{x}_j)}} \nonumber \\ &- \mu \sum_{i=1}^{N} \sum_{k=1}^{K} m_{ik} \mathrm{log}\left(m_{ik}\right)
\end{align}
Getting the correct membership vector then is equivalent to solving this constrained optimization problem:
\begin{align}
\label{eq:optimization}
&\text{min}_{\substack{\mathbf{m}_1,\dots,\mathbf{m}_N}} \hat{J}_{CS}^{REG}(\mathbf{m}_1,\dots,\mathbf{m}_N) \quad \nonumber \\ &\text{subject to } \mathbf{m}_j^\mathrm{T}\mathbf{1} - \mathbf{1} = 0, \quad j = 1,\dots ,N 
\end{align}

This can be solved using Lagrange multipliers and a stochastic gradient descent scheme to compute optimum values of $\mathbf{m}$. In the second step, a kernel based regression technique that uses a training set of pixels and fits a function to it, by minimizing the representational error, is used to generate the downscaled estimates. Ridge regression is a parametric regression technique that adds a scaled regularizing term to the cost function to to increase generalization. The cost function for ridge regression is

\begin{equation}
\mathcal{E}\left( \mathbf{w},\mathbf{x}\right)   = \frac{1}{2} \sum_i (y_i - w^{\mathbf{T}}\mathbf{x}_i)^2 + \frac{1}{2} \mu \|\mathbf{w}\|^2
\end{equation}

The weights can be calculated by differentiating the error cost function with respect to the weights and setting it to zero. If this computation was carried out in a Reproducing Kernel Hilbert Space (RKHS), then the inner-products can be replaced with a kernel evaluation. Let $\mathcal{H}$ be a Hilbert space with an inner-product metric $<\cdot,\cdot>_{\mathcal{H}}$. Then according to the representer theorem, a kernel function $\kappa(\mathbf{x},\mathbf{y})$ exists on $\mathbb{R}^\mathrm{N} \times \mathbb{R}^\mathrm{N}$ such that $<\mathbf{x},\mathbf{y}>_{\mathcal{H}} = \kappa(\mathbf{x},\mathbf{y})$. Now, if $\Phi:\mathbb{R}^\mathrm{N} \rightarrow \mathbb{R}^\mathrm{N}$ is a mapping that transforms the feature vector in the original vector space to $\mathcal{H}$, then the weights can de redefined as,

\begin{equation}
\mathbf{w} = (\mu \mathbf{I}_D + \Phi \Phi^\mathrm{T})^{-1} \Phi\mathbf{y}
\end{equation}

Then, the estimated value of y for a new data-point $\mathbf{x'}$ is,
\begin{equation}
\hat{y} = \mathbf{w}^\mathrm{T} \Phi(\mathbf{x}') = \mathbf{y}(\mu \mathbf{I}_N + \mathbf{\mathrm{K}} )^{-1} \kappa(\mathbf{x},\mathbf{x}')
\end{equation}

where $\mathbf{\mathrm{K}}$ is the Gram matrix of inner products of all the training data points. This does not address the constant that must be present in the regression. To solve this problem, the feature vector is augmented by adding a constant feature 1 to all samples.

\section{Methodology}

\subsection{Multiscale synthetic dataset}

The proposed algorithm for disaggregation was tested using data generated by a simulation framework consisting of the Land Surface Process (LSP) model and the Decision Support System for Agrotechnology Tranfer (DSSAT) model, described in \cite{Nagarajan2012}. A $50\times50$ km$^2$ region, equivalent to 25 SMAP pixels, was chosen in North Central Florida for the simulations.

Based upon land cover information at 200 m, contiguous, homogeneous regions of sweet-corn and cotton were identified. A realization of the LSP-DSSAT model was used to simulate LST, LAI, and PPT at the centroid of each homogeneous region, using the corresponding crop module within DSSAT. The model simulations were performed using the 200 m forcings at the centroid. Linear averaging is typically sufficient to illustrate the effects of resolution degradation \cite{Crow2003}. The model simulations at 200 m were spatially averaged to obtain PPT, LST, LAI, and SM at 1 and 10 km.  To simulate rain-fed systems, all the water input from both precipitation and irrigation were combined together, and the ``PPT" in this study represents these combined values, representing a rainfed system.

%======================================
%\subsection{Microwave Brightness (MB) model for T$_{\textrm{B}}$ estimation from \textit{In-Situ} SM}
%\label{p2p3:mbmodel}
%======================================
T$_{\textrm{B}}^\mathrm{1 km}$ measurements, were generated from $\mathrm{SM}_i^\mathrm{1 km}$ for the entire area, using the widely used $\tau-\omega$ model~\cite{Liu2013}. 10 \% of the $\mathrm{T_B}$ measurements were used for training. The rest was used as "truth" for validating the downscaled $\mathrm{T_B}$. An alternative and better approach is to convert the downscaled $\mathrm{T_B}$ into SM using an inverse model or assimilation~\cite{Reichle2001}, since SM is the key variable required at high resolution. However, that detracts from the focus of this study by adding an extra layer of complexity. A vegetated surface is modeled as a single isothermal layer of vegetation with diffuse boundaries \cite{Nagarajan2012}. The soil medium is assumed to be non-isothermal, semi-infinite layered dielectric medium with a rough surface at the upper boundary. The soil moisture and temperature profiles, the leaf/ear biomass, vegetation water content, and plant height provided by the LSP-DSSAT model were used by the MB model to estimate brightness temperature at L-band. Using a zeroth-order radiative transfer approach, the total T$_{\textrm{B}}$ of a terrain is the sum of contributions from soil, vegetation, and from sky. In this study, similar to \cite{Chakrabarti2014} the $\theta$ was set to $50^o$, the $T_{sky}$ was set to 5K, $r_p$ was obtained by integrating the bistatic scattering coefficients from the IEM model, the RMS height was 0.62~cm and, the correlation length was 8.72~cm~\cite{Liu2013}. Other model parameters are set similar to~\cite{Nagarajan2012}.
%-----------------------------------
\begin{figure}[t]
\centering
\includegraphics[width=3.5in, trim= 3cm 0cm 0cm 1.8cm, clip=true ]{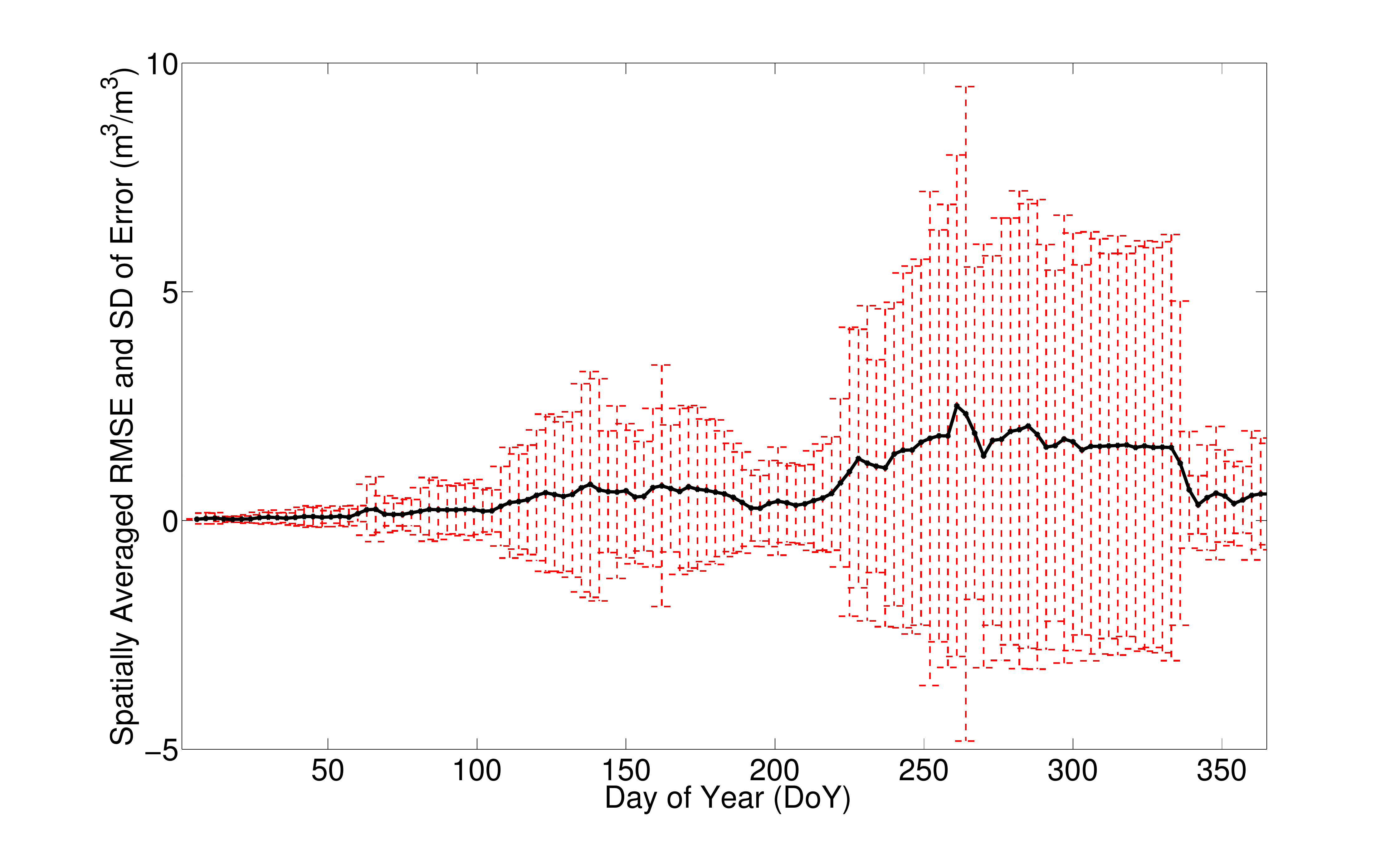}%
\caption{Root Mean Square Error (RMSE) and Standard Deviation (SD) between true $\mathrm{T_B}$ and downscaled $\mathrm{T_B}$ at 1~km. }
\label{fig:RMSE}
\end{figure}
%-------------------------------------
\subsection{Implementation of Disaggregation Framework based on Self-Regularized Regressive Models}
The simulation period, from Jan 1 (DoY 1) to Dec 31 (DoY 365), 2007, consisted of two growing seasons of sweet corn and one season of cotton. The LST, PPT, and LAI observations at 1 km were obtained by adding noise to account for satellite observation errors, instrument measurement errors, and micro-meteorological variability, following \cite{Huang2008b}. Errors with zero mean and standard deviations of 5K, 1 mm, and 0.1 for LST, PPT, and LAI, respectively, were added to the values at 10 k.

The SRRM method uses LST, 3-day PPT, LAI, LC at 1 km and $\mathrm{T_B}$ at 10 km every 3 days as input. In the first step, the field is clustered using the inputs at 1~km and the x and y coordinates of each pixel scaled to a range of 0 and 1. This step of the algorithm uses two parameters - the number of clusters, $\mathrm{n}$ and a regularization constant, $\mu$. Both the number of clusters and the regularization constant is determined by cross-validating against the absolute mean error in $\mathrm{T_B}$ at the end of the second step for each day. The optimal number of iterations that produces a usable clustering result is determined by evaluating the root mean square error (RMSE) for Day 222,  characterized by maximum input heterogeneity, in disaggregated $\mathrm{T_B}$ after every iteration, for upto 200 iterations. At the end of this step, each pixel has a vector of $\mathrm{n}$ numbers, $(m_1,m_2,\dots,m_N)$ that sum upto 1 describing its membership to each of the $\mathrm{n}$ clusters.
\begin{figure}[t]
\centering
\includegraphics[width=3in, trim= 2cm 3cm 7cm 1.5cm, clip=true ]{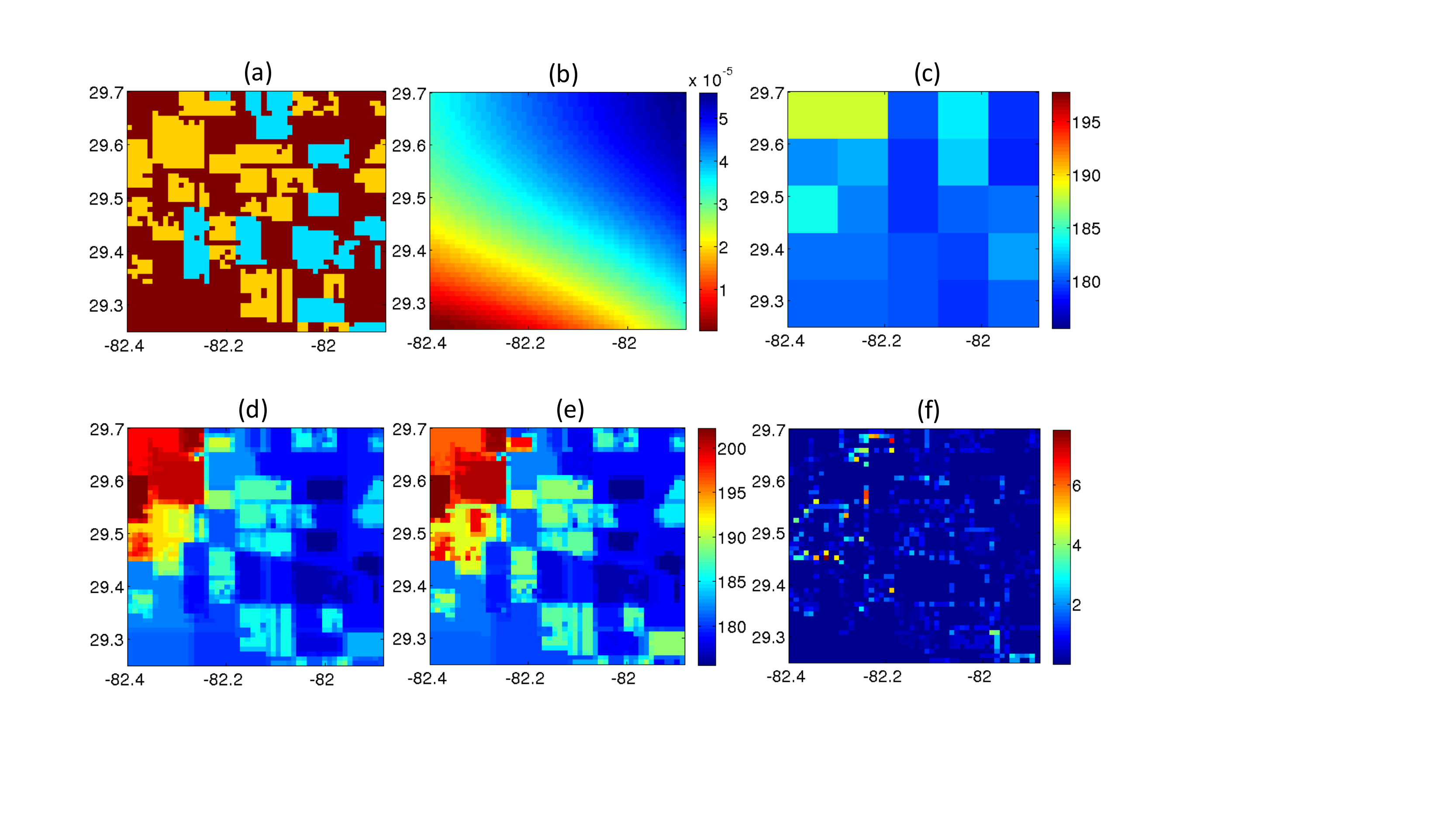}%
\caption{Day 195 - (a) LC at 1 km (blue represents baresoil, yellow represents sweet-corn and cyan represents cotton), (b) PPT at 1 km (in $\mathrm{m^3/m^3}$), (c) $\mathrm{T_B}$ at 10~km, (d) True $\mathrm{T_B}$ at 1~km (in K), (e) Downscaled $\mathrm{T_B}$ at 1~km (in K), (f) Absolute Difference Plot between True and Downscaled $\mathrm{T_B}$ (in K). X-axis denotes longitudes ($^{\circ}$W) and Y-axis denotes latitudes ($^{\circ}$N) for every plot. }
\label{fig:Day65}
\end{figure}
In the second step, $\mathrm{n}$ models, $\hat{f}_1, \hat{f}_2,\dots,\hat{f}_N$ are developed using LST, 3-day PPT, LAI, LC, $\mathrm{T_B}$ at 1 km and $\mathrm{T_B}$ at 10 km as inputs to the regularized kernel regression algorithm described in Section~\ref{Sec:Theory}, using 10\% of the pixels that make up the field. The hard membership of each pixel, $i$, for model development purposes is determined by the maximum value in its membership vector, $\mathbf{m}^i=(m_1^i,m_2^i,\dots,m_N^i)$. The disaggregated value of $\mathrm{T_B}$ is computed for each point in the test, represented as a vector, $\mathbf{x'}_i = (\mathrm{LST}_i^\mathrm{1 km}, \mathrm{PPT}_i^\mathrm{1 km}, \mathrm{LAI}_i^\mathrm{1 km}, \mathrm{LC}_i^\mathrm{1 km}, \mathrm{T_B}^\mathrm{1 km})$ by,

%****************************************************************************************
\begin{equation}
\mathrm{\mathrm{T_B}}_i^\mathrm{1 km} = \mathbf{m}^\mathrm{T}\cdot \left(\hat{f}_1(\mathbf{x'}_i), \hat{f}_2(\mathbf{x'}_i),\dots,\hat{f}_N(\mathbf{x'}_i)\right)
\end{equation}
%****************************************************************************************
The SRRM method is evaluated by studying the RMSE and standard deviation of the errors over the entire season. Moreover, the downscaled $\mathrm{T_B}$ is plotted versus the true $\mathrm{T_B}$, for each land-cover. To evaluate how close the density function of the downscaled estimates is to the density function of the true SM, the KL-Divergence (KLD) between the density of the estimated observations and the true $\mathrm{T_B}$ is calculated for different LC's over the season. The PPT, LC, low resolution $\mathrm{T_B}$, true $\mathrm{T_B}$ and downscaled $\mathrm{T_B}$ are examined for representative days when the crop variability and micro-meterological variability was highest. Quantitative analyses of spatial variations in $\mathrm{T_B}$ observed under dynamic vegetation and heterogeneous land cover conditions provide an index of dynamic errors that can be expected. 

%****************************************************************************************
\section{Results}
\begin{figure}[t]
\centering
\includegraphics[width=0.48\textwidth, trim= 0.5cm 6cm 1cm 3cm, clip=true ]{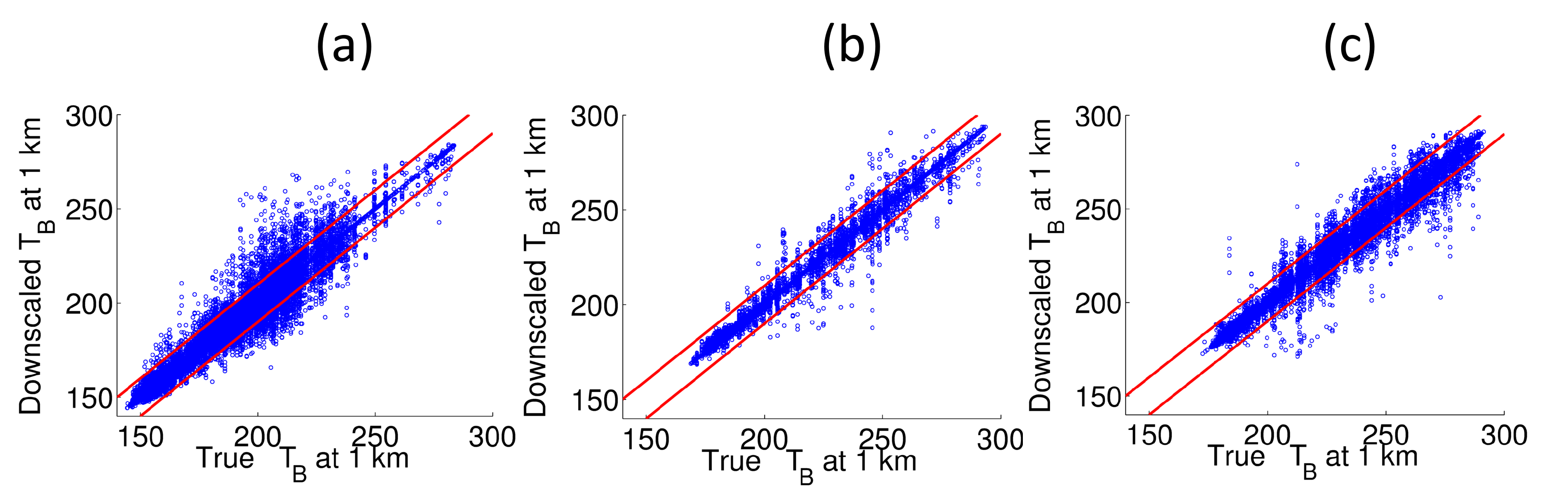}%
\caption{Disaggregated $\mathrm{T_B}$ vs. True $\mathrm{T_B}$ at 1 km during the whole season for (a)baresoil pixels (b)corn pixels, and (c)cotton pixels. Lines corresponding to 10~K error in $\mathrm{T_B}$ are shown for each plot. }
\label{fig:LandCover}
\end{figure}

The RMSE and SD of error between true and downscaled $\mathrm{T_B}$ at 1~km is shown in Figure~\ref{fig:RMSE}. The standard deviation increases from below 3~K to almost 7~K during the vegetated season. This is due to a large difference in errors with most of the errors originating from pixels on the field boundaries. A Z-test was performed which showed that the errors lie under 10~K overall for more than 95\% of the pixels. They were found to be the highest during the growing season for sweet-corn and cotton, and the late season baresoil land cover after harvest, due to residual vegetation after harvest. Figure~\ref{fig:LandCover} shows the errors for each land-cover. The errors are high, during the vegetated season, for baresoil land-cover due to sub-pixel heterogeneities. Pixels at 1~km are classified as baresoil when less than half of the area is cultivated. These impure pixels contribute to the high error. This is also evident in Figure~\ref{fig:Day65}, which shows the LC, PPT, $\mathrm{T_B}$ at 10 km, true $\mathrm{T_B}$, downscaled $\mathrm{T_B}$, and the absolute difference between true and downscaled $\mathrm{T_B}$ at 1~km for Day~195. The absolute difference plot indicates that the maximum errors occur at the edges of the fields which confirms the impure pixel hypothesis. The true and downscaled $\mathrm{T_B}$ are very similar which shows that the algorithm can perform well during heterogeneous micro-meteorological conditions.

The RMSE, SD and KL-Divergence (KLD) over the whole season is shown in Table~\ref{tab:pritable}. Baresoil pixels at the end of the season affected by remnant crops and impure pixels have a higher KLD, but very close to 0. Vegetated pixels at 1~km have a higher KLD as well. Vegetated pixels with baresoil impurities contribute to these errors. As expected the RMSE is lowest during early season baresoil, when the downscaled estimates are almost perfect. The RMSE's and SD's increase during the growing season, which is expected due to the complexity introduced by different land-cover types.
\begin{table}
\renewcommand{\arraystretch}{1.5} 
\centering
\caption{RMSE, SD, and KL divergence over the 50$\times$50 km$^2$ region for the downscaled estimates of SM obtained at 1~km using the PRI AND UT methods. (\textbf{A} - Baresoil pixels with vegetated sub-pixels at 250 m till DoY 332, \textbf{B} - Baresoil pixels after DoY 332 and \textbf{C} - Baresoil pixels without any  vegetated sub-pixels at 250 m till DoY 332)}
\label{tab:pritable}
\begin{tabular}{c|c|c|c}
\hline
\multicolumn{1}{c|}{Land Cover} & $\mathrm{KLD_{SRRM}}$  & $\mathrm{RMSE_{SRRM}}$ & $\mathrm{SD_{SRRM}}$\\
\hline

Corn & $1.8615\times 10^{-17}$ &4.1 K& 2.5 K\\
Cotton &$2.4828\times 10^{-04}$ &3.7 K& 3.4 K\\
Baresoil$^\mathrm{A}$ & $5.6222\times 10^{-5}$ &7.7 K& 1.51 K\\
Baresoil$^\mathrm{B}$& $5.628\times 10^{-6}$ &8.1 K& 1.8 K\\
Baresoil$^\mathrm{C}$ & $2.5948\times 10^{-6}$ &1.2 K& 0.9 K\\
\end{tabular}
\end{table}
\section{Conclusion}
%****************************************************************************************
In this study, a downscaling methodology for $\mathrm{T_B}$, based upon SRRM models, was implemented and evaluated to work with an RMSE of 5.76~K with standard deviation of 2.8~k, during the vegetated season, and an RMSE of 1.2~K with a standard deviation of 0.9~K during periods of no vegetation. Thus this algorithm will be particularly useful during heterogeneous LC and micro-meteorological conditions, with limited set of training observations available. It is envisioned that this algorithm will be applied using the SM observations available from the soon to be launched NASA-Soil Moisture Active Passive (SMAP) mission in conjunction with remotely sensed LST, LAI and PPT data from other satellites in the Earth Observing System. Future scope of research includes data-fusion of SMOS observations of $\mathrm{T_B}$ at multiple angles with SMAP $\mathrm{T_B}$ to increase the information content in the low resolution product for better downscaling accuracies, and using variational data-assimilation to calculate SM from the downscaled $\mathrm{T_B}$.

% trigger a \newpage just before the given reference
% number - used to balance the columns on the last page
% adjust value as needed - may need to be readjusted if
% the document is modified later
%\IEEEtriggeratref{8}
% The "triggered" command can be changed if desired:
%\IEEEtriggercmd{\enlargethispage{-5in}}

% references section

% can use a bibliography generated by BibTeX as a .bbl file
% BibTeX documentation can be easily obtained at:
% http://www.ctan.org/tex-archive/biblio/bibtex/contrib/doc/
% The IEEEtran BibTeX style support page is at:
% http://www.michaelshell.org/tex/ieeetran/bibtex/
%\bibliographystyle{IEEEtran}
% argument is your BibTeX string definitions and bibliography database(s)
%\bibliography{IEEEabrv,../bib/paper}
%
% <OR> manually copy in the resultant .bbl file
% set second argument of \begin to the number of references
% (used to reserve space for the reference number labels box)
%----------------------------------------------------------------------------------------
%	BIBLIOGRAPHY
%----------------------------------------------------------------------------------------

\renewcommand{\baselinestretch}{1.0}
\footnotesize
\bibliographystyle{IEEEtran}
\bibliography{monsivaisIEEE}

% Generated by IEEEtran.bst, version: 1.12 (2007/01/11)
\begin{thebibliography}{10}
\providecommand{\url}[1]{#1}
\csname url@samestyle\endcsname
\providecommand{\newblock}{\relax}
\providecommand{\bibinfo}[2]{#2}
\providecommand{\BIBentrySTDinterwordspacing}{\spaceskip=0pt\relax}
\providecommand{\BIBentryALTinterwordstretchfactor}{4}
\providecommand{\BIBentryALTinterwordspacing}{\spaceskip=\fontdimen2\font plus
\BIBentryALTinterwordstretchfactor\fontdimen3\font minus
  \fontdimen4\font\relax}
\providecommand{\BIBforeignlanguage}[2]{{%
\expandafter\ifx\csname l@#1\endcsname\relax
\typeout{** WARNING: IEEEtran.bst: No hyphenation pattern has been}%
\typeout{** loaded for the language `#1'. Using the pattern for}%
\typeout{** the default language instead.}%
\else
\language=\csname l@#1\endcsname
\fi
#2}}
\providecommand{\BIBdecl}{\relax}
\BIBdecl

\bibitem{Chakrabarti2014}
S.~Chakrabarti, T.~Bongiovanni, J.~Judge, K.~Nagarajan, and J.~C. Principe,
  ``Downscaling satellite-based soil moisture in heterogeneous regions using
  high-resolution remote sensing products and information theory: A synthetic
  study,'' \emph{{\it IEEE Trans. Geosci. Remote Sensing}}, vol.~53, no.~1, pp.
  85--101, 2014.

\bibitem{Chakrabarti2013}
S.~Chakrabarti, T.~Bongiovanni, J.~Judge, L.~Zotarelli, and C.~Bayer,
  ``Assimilation of smos soil moisture for quantifying drought impacts on crop
  yield in agricultural regions,'' \emph{{\it IEEE J. Sel. Topics Appl. Earth
  Observ}}, vol.~7, no.~9, pp. 3867--3879, 2013.

\bibitem{Merlin2013}
O.~Merlin, M.~Escorihuela, M.~Mayoral, O.~Hagolle, A.~A. Bitar, and Y.~Kerr,
  ``Self-calibrated evaporation-based disaggregation of smos soil moisture: an
  evaluation study at 3 km and 100 m resolution in catalunya spain,''
  \emph{{\it Remote Sens. Env.}}, vol. 130, no.~0, pp. 25--38, 2013.

\bibitem{Piles2012}
M.~Piles, M.~Vall-llossera, L.Laguna, and A.~Camps, ``A downscaling approach to
  combine smos multi-angular and full-polarimetric observations with mdis
  vis/ir data into high resolution soil moisture maps,'' in \emph{{IEEE Int.
  Geosc. and Rem. Sens. Symposium}}, vol.~1.\hskip 1em plus 0.5em minus
  0.4em\relax Proc. {IGARSS 2012}, 2012, pp. 1247--1250.

\bibitem{Das2014}
N.~Das, D.~Entekhabi, and E.~Njoku, ``Tests of the smap combined radar and
  radiometer algorithm using airborne field campaign observations and simulated
  data,'' \emph{{\it IEEE Trans. Geosci. Remote Sensing}}, vol.~52, no.~4, pp.
  2018--2028, 2014.

\bibitem{YongQian2011}
Y.~Wang, J.~Shi, L.~Jiang, J.~Du, and B.~Tian, ``The development of an
  algorithm to enhance and match the resolution of satellite measurements from
  amsr-e,'' \emph{{\it Science China: Earth Sciences}}, vol.~54, no.~3, pp.
  410--419, 2011.

\bibitem{Gambardella2008}
A.~Gambardella and M.~Migliaccio, ``On the superresolution of microwave
  scanning radiometer measurements,'' \emph{{\it IEEE Geosci. and Remote
  Sensing Letters}}, vol.~5, no.~4, pp. 796--800, 2008.

\bibitem{Lenti2014}
F.~Lenti, F.~Nunziata, C.~Estatico, and M.~Migliaccio, ``On the spatial
  resolution enhancement of microwave radiometer data in banach spaces,''
  \emph{{\it IEEE Trans. Geosci. Remote Sensing}}, vol.~52, no.~3, pp.
  1834--1842, 2014.

\bibitem{Nagarajan2012}
K.~Nagarajan and J.~Judge, ``Spatial scaling and variability of soil moisture
  over heterogenous land cover and dynamic vegetation conditions,'' \emph{{\it
  IEEE Geosci. and Remote Sensing Letters}}, vol.~10, no.~4, pp. 880--884,
  2013.

\bibitem{Jenssen2005}
R.~Jennsen, D.~Erdogmus, K.~Hild, J.~Principe, and T.~Eltoft, ``Optimizing the
  cauchy-schwarz pdf distance for information theoretic, non-paremetric
  clustering,'' in \emph{{Proceedings of the 5th international conference on
  Energy Minimization Methods in Computer Vision and Pattern Recognition}},
  vol.~1.\hskip 1em plus 0.5em minus 0.4em\relax Proc. {EMMCVPR 2005}, 2005,
  pp. 34--45.

\bibitem{Crow2003}
W.~Crow and E.~Wood, ``The assimilation of remotely sensed soil brightness
  temperature imagery into a land surface model using {Ensemble Kalman}
  filtering: a case study based on {ESTAR} measurements during {SGP97},''
  \emph{{\it Adv. in Water Res.}}, vol.~26, no.~2, pp. 137--149, 2003.

\bibitem{Liu2013}
P.~Liu, R.~DeRoo, A.~England, and J.~Judge, ``Impact of moisture distribution
  within the sensing depth on l- and c-band emission in sandy soils,''
  \emph{IEEE J. Sel. Topics Appl. Earth Observ.}, vol.~6, no.~2, pp. 887--899,
  2013.

\bibitem{Reichle2001}
R.~H. Reichle, D.~Entekhabi, and D.~B. McLaughlin, ``Downscaling of radio
  brightness measurements for soil moisture estimation: {A} four-dimensional
  variational data assimilation approach,'' \emph{{\it Water Res. Research}},
  vol.~37, no.~9, pp. 2353--2364, 2001.

\bibitem{Huang2008b}
C.~Huang, X.~Li, and L.~Lu, ``Retrieving soil temperature profile by
  assimilating {MODIS} {LST} products with ensemble {K}alman filter,''
  \emph{{\it Remote Sens. Env.}}, vol. 112, pp. 1320--1336, 2008.

\end{thebibliography}
%----------------------------------------------------------------------------------------
% that's all folks
\end{document}